\documentclass{article}
\usepackage{spconf,amsmath,epsfig}
\usepackage{bm}
\usepackage{color}
\usepackage{subfig}
\usepackage{amssymb}
\usepackage{paralist}
\usepackage{amsfonts}
\usepackage{booktabs}
\usepackage{enumerate}
\usepackage{multirow}
\usepackage{threeparttable}
\let\OLDthebibliography\thebibliography
\renewcommand\thebibliography[1]{
  \OLDthebibliography{#1}
  \setlength{\parskip}{0pt}
  \setlength{\itemsep}{0pt plus 0.3ex}
}

\begin{document}\sloppy

% Example definitions.
% --------------------
\def\x{{\mathbf x}}
\def\L{{\cal L}}

% Title.
% ------
\title{Transfering Low-Frequency Features for Domain Adaptation}
%
% Single address.
% ---------------
\name{Zhaowen Li\textsuperscript{1,2}, Xu Zhao\textsuperscript{1}, Chaoyang Zhao\textsuperscript{1,3}, Ming Tang\textsuperscript{1} and Jinqiao Wang\textsuperscript{1,2}}
%Address and e-mail should NOT be added in the submission paper. They should be present only in the camera ready paper. 
\address{\textsuperscript{1} National Laboratory of Pattern Recognition, Institute of Automation, \\
	Chinese Academy of Sciences, Beijing, China \\
	\textsuperscript{2} School of Artificial Intelligence, University of Chinese Academy of Sciences, \\
	Beijing, China\\
	\textsuperscript{3} Development Research Institute of Guangzhou Smart City\\
	{\tt\small {\{zhaowen.li,xu.zhao,chaoyang.zhao,tangm,jqwang\}@nlpr.ia.ac.cn}}
}

\maketitle

\begin{abstract}
Previous unsupervised domain adaptation methods did not handle the cross-domain problem from the perspective of frequency for computer vision. The images or feature maps of different domains can be decomposed into the low-frequency component and high-frequency component. This paper proposes the assumption that low-frequency information is more domain-invariant while the high-frequency information contains domain-related information. Hence, we introduce an approach, named low-frequency module (LFM), to extract domain-invariant feature representations. The LFM is constructed with the digital Gaussian low-pass filter. Our method is easy to implement and introduces no extra hyperparameter. We design two effective ways to utilize the LFM for domain adaptation, and our method is complementary to other existing methods and formulated as a plug-and-play unit that can be combined with these methods. Experimental results demonstrate that our LFM outperforms state-of-the-art methods for various computer vision tasks, including image classification and object detection.
\end{abstract}
\begin{keywords}
domain adaptation, unsupervised, frequency learning
\end{keywords}
\section{Introduction}
\label{intro}

%As a feature extraction mechanism, convolutional neural networks (CNNs) have shown good performance on various supervised computer vision tasks, such as image recognition \cite{2015ImageNet,he2016deep,huang2017densely},  object detection \cite{girshick2015fast,ren2017faster,lin2020focal}, and instance segmentation \cite{he2020mask,chen2018encoder}. Massive labeled training data are a key to learning a successful deep model. Unfortunately, collecting, annotating, and managing large scale datasets is expensive and laborious ordeal reserved for researchers. 

Unsupervised domain adaptation (UDA) methods can transfer a learner for the target domain data while manual annotations are only provided in source domain data. The principal idea of UDA methods is to mitigate the domain shift in data distributions. For the UDA image classification task, some previous work \cite{long2015learning,long2017deep} minimize the domain-discrepancy to obtain domain-invariant feature representations in convolutional neural networks (CNNs), where the domain-discrepancy is measured by Maximum Mean Discrepancy (MMD) \cite{long2015learning} or Joint MMD (JMMD) \cite{long2017deep}. Another popular idea for UDA is to adopt an adversarial learning method to obtain domain-invariant features. RevGrad \cite{ganin2015unsupervised} is a representative of these adversarial learning methods by back-propagating the reverse gradients of the domain classifier. These methods mainly focus on learning a global domain shift, aligning the global source and target distributions without considering the category information in both domains. Recently, some researchers considered that making use of pseudo label can help the network to better align domain-invariant features. For example, based on MMD, CAN \cite{kang2019contrastive} proposed a contrastive adaptation network, which optimizes contrastive domain discrepancy explicitly modeling the intra-class domain discrepancy and the inter-class domain discrepancy. Additionally, there are some methods that are specifically designed for object detection. The authors of \cite{chen2018domain} presented two domain adaptation components, image-level adaptation and instance-level adaptation. They adopt domain adversarial approach using a discriminator for each component. The similar motivation was used to align feature representation across domains on enlarged positive regions \cite{zhu2019adapting}. Mean Teacher with object relations \cite{cai2019exploring} was also considered, which addressed the adaptive detection from the viewpoint of graph-structured consistency. However, the above methods do not handle the domain adaptation problem from the perspective of frequency.

\begin{figure}
    \small
    \centering
    \includegraphics[width=0.4\linewidth]{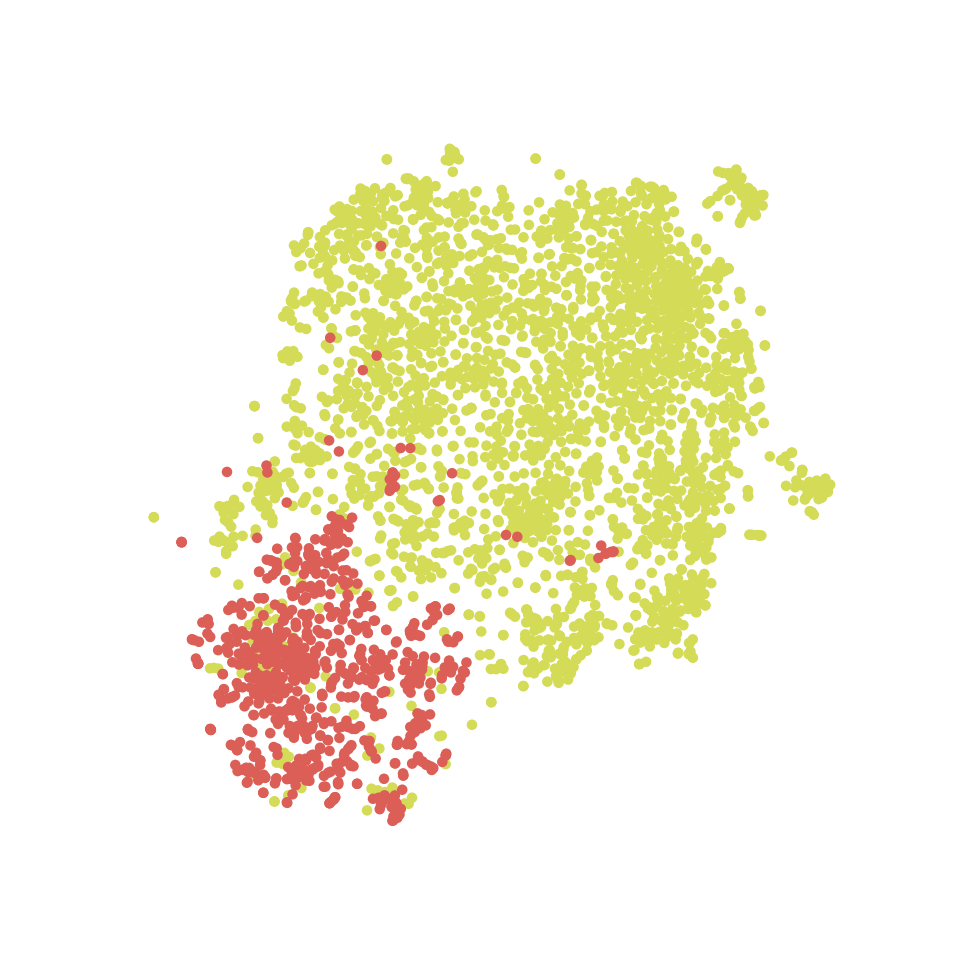}
    \includegraphics[width=0.4\linewidth]{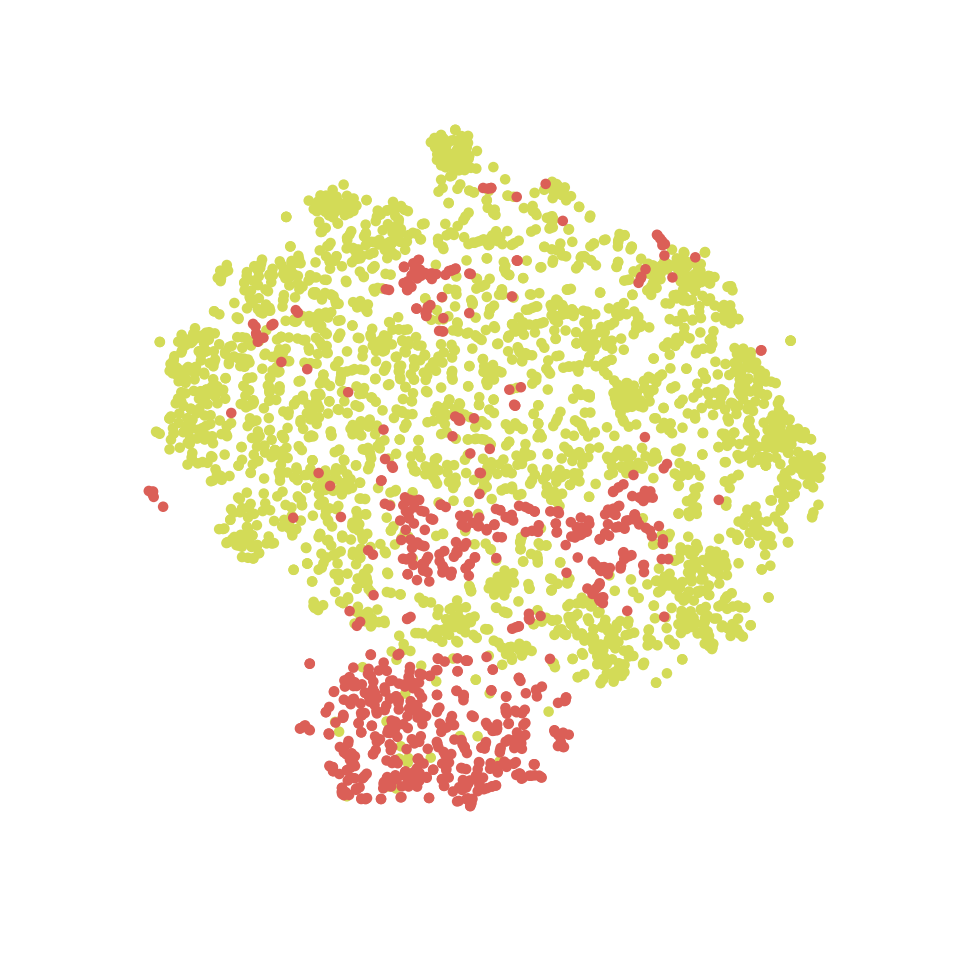}
    \vspace{-4mm}
    \caption{ t-SNE \cite{maaten2008visualizing} visualization of the distribution of original image data and low-frequency data in the \textbf{Amazon (A)} domain and \textbf{Webcam (W)} domain of the Office-31 dataset \cite{saenko2010adapting}. \textbf{Left}: t-SNE of original image data. \textbf{Right}: Low-frequency data. The \textbf{A} domain consists of 2817 images (yellow point) and \textbf{W} domain consists of 795 images (red point). }
    \vspace{-4mm}
    \label{tsne}
\end{figure}

It is well known \cite{devalois1990spatial,2008Digital,chen2019drop} that a single natural image or feature map can be decomposed into a low-frequency component that describes the smoothly changing structure, and a high-frequency component that describes the rapidly changing fine details. This paper proposes an assumption that the low-frequency information of the same class in different domains has domain-invariant characteristics. To better present this statement, in Fig.~\ref{tsne}, we visualize the distribution of original image data and low-frequency data obtained from original data processed by the digital Gaussian low-pass filter \cite{2008Digital}. Fig.~\ref{tsne} illustrates the data distributions of the \textbf{W} domain and \textbf{A} domain of the Office-31 dataset. The \textbf{W} domain and \textbf{A} domain consist of the images of the same classes. On the left side of Fig.~\ref{tsne}, the domain-discrepancy of the two distinct domains using the whole-frequency information is large. In contrast, on the right, the domain-discrepancy of the two domains utilizing the low-frequency information is reduced. Hence, it is reasonable to assume that the low-frequency information is more domain-invariant than the whole-frequency information contained in datasets. Meanwhile, the high-frequency information suppressed by the digital Gaussian low-pass filter contains domain-related information and easily affects the alignment of the data distribution. 

To improve the generalization performance of the network, in this paper, we propose a simple yet effective method called low-frequency module (LFM). The LFM is constructed with the digital Gaussian low-pass filter. It can enhance the generalization performance of models by utilizing the inherent low-frequency information of feature maps. This method is straightforward to implement, and introduces no extra hyperparameter. Experimental results on various benchmarks demonstrate the effectiveness of our method. To summarize, our contributions are as follows:
\begin{enumerate}
    \item We propose a novel domain adaptation technique called LFM. We show that the LFM can help CNNs achieve better generalization performance by utilizing the inherent low-frequency information of feature maps in various domain adaptation tasks.
    \item We propose two different ways to utilize the LFM and validate the effectiveness of our method on standard benchmarks for different tasks, such as image classification and object detection.
    \item Our method achieves state-of-the-art performance on VisDA-2017 \cite{peng2017visda} and Cityscapes \cite{cordts2016the} to FoggyCityscapes \cite{sakaridis2018semantic}. 
\end{enumerate}

\section{Approach}
In this section, we first introduce the domain adaptation problem and provide a discussion about the characteristics of the low-frequency information. Then, we reveal the relationship between the domain adaptation problem and the low-frequency information. Finally, we analyze our proposed LFM and how to use the LFM.

\subsection{Domain Adaptation}
\label{ada}
The domain adaptation problem can be viewed as aligning part or global feature representations in the learned feature extractor. For example, we consider classification tasks where $X$ is the input space and $Y$ is the set of possible labels. In fact, we have two different distributions over $X \times Y$, called the source and the target domains. An UDA learning algorithm is provided with labeled samples drawn from the source domain, and unlabeled samples drawn from the target domain. The purpose of the UDA algorithm is to make the samples of the same annotation in different domains eventually outputs similar feature representations eventually. 

Researchers \cite{long2015learning,long2017deep,kang2019contrastive} take the \emph{source-finetune} method as the basic method of domain adaptation. The \emph{source-finetune} makes the CNN model directly train on the source domain data and predict on the target domain data. Taking the classification task with ResNet \cite{he2016deep} as an example, at training time, the optimization process is given with Eq (\ref{eq1}), where $\bm {\hat{\theta_f}, \hat{\theta_g}}$ are optimized parameters trained on the source domain $X_S \times Y_S$. The $\bm \theta_g$ represents the parameters of linear classification layer $\bm g(\cdot)$, and the $\bm \theta_f$ is the parameter of CNN encoder $\bm{f(\cdot)} $. The the global average pooling layer is $\bm p(\cdot)$, and $\bm L(\cdot)$ is the cross entropy loss. 
\begin{equation}
\begin{split}
    \hat{\theta_f}, \hat{\theta_g} = {\rm arg min_{\theta_f, \theta_g}} \left[ L\left( g\left(  p \left(  f(X_S,\theta_f), 
    \theta_g \right) \right),Y_S\right)  \right]
    \label{eq1}
\end{split}
\end{equation}

The existing UDA methods adopt various methods to make the parameters $\bm {\hat{\theta_f}, \hat{\theta_g}}$ adapt to the target domain. However, these methods do not deal with the domain adaptation from the perspective of frequency.

\subsection{Low-frequency Information}
According to \cite{devalois1990spatial,2008Digital,chen2019drop}, a single natural image or feature map can be decomposed into a low-frequency component that describes the the structure information and a high-frequency component that describes the rapidly changing fine details and noise. We propose an assumption that that the low-frequency components are the key information for cross-domain tasks for the following reasons: 1) Inspired by the idea above, \emph{the low-frequency information of the image or feature map reflects shape information}. Although the overall data distribution is different in different domains, the same objects of the same class label have similar shapes. We hypothesize that the shape information can represent the intrinsic characteristics of the objects.  
2) In addition, from the perspective of the signal process, \emph{the low-frequency information represents the main components of a two-dimensional signal} \cite{devalois1990spatial}. Utilizing the low-frequency information does not change the main components. 
3) Moreover, as shown in Fig.~\ref{tsne}, we find that \emph{the domain-discrepancy between the \textbf{W} and \textbf{A} domain is reduced by the digital Gaussian low-pass filter}. We conclude that the low-frequency information is more domain-invariant than the whole information. The low-pass filter passes the low-frequency information while suppressing the high-frequency information. Hence, we argue that the suppressed high-frequency information contains domain-related information. Simultaneously, we argue that part of the reason for the domain-discrepancy is the significant differences in high-frequency information between different domains.
4) Notably, the experimental results also demonstrate our assumption in Experiment~\ref{low-exp}.

In conclusion, it is reasonable to assume that the low-frequency information is more domain-invariant and suitable for domain adaptation tasks while high-frequency information may harm the generalization performance and stability of the model.

\subsection{The Low-Frequency Module for Domain Adaptation}

In this section, we propose a low-frequency module (LFM) to help the network align low-frequency feature representations. 

\begin{figure}
    \small
    \centering
    
    \includegraphics[width=1.0\linewidth]{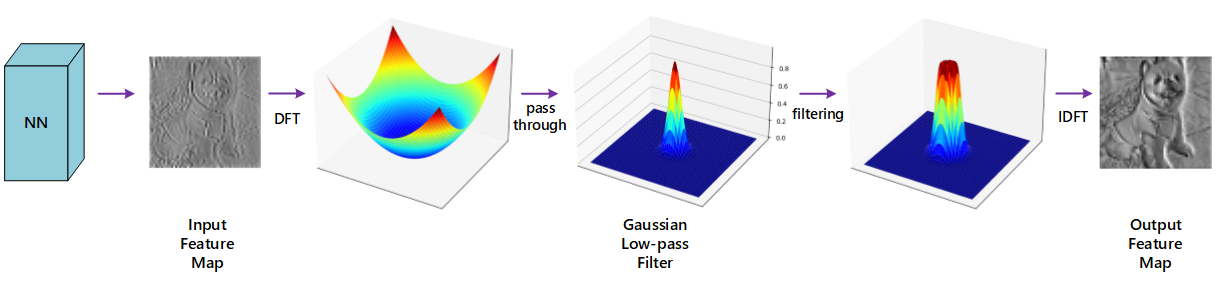}
    \vspace{-6mm}
    \caption{The processing procedure of LFM. The input spatial feature map obtained by the neural network (NN)  is  converted  to  the  distribution  of  frequency by Discrete Fourier Transform (DFT) \cite{2008Digital}.}
    \label{3dguss}
    \vspace{-5mm}
\end{figure}

\subsubsection{The LFM}

\emph{The LFM is essentially a digital low-pass filter}. The principle of the low-pass filter is to pass the low-frequency information while suppressing the high-frequency information for two-dimensional discrete signal \cite{2008Digital}. In this paper, the digital low-pass filter adopts the Gaussian low-pass filter \cite{2008Digital} with kernel $m \times m$. Because the Gaussian low-pass filter has no ringing \cite{2008Digital}, it makes the quality of extracted low-frequency information of the whole information better than other low-pass filters, such as the ideal low-pass filter \cite{2008Digital}. 

As shown in Fig.~\ref{3dguss}, the spatial feature map obtained by neural network (NN) is converted to the distribution of frequency by DFT.  Assuming the distribution of input feature map in the frequency domain, the high-frequency information of the output is filtered out when the input passes through the Gaussian low-pass filter. Finally, the output feature map is obtained by Inverse Discrete Fourier Transform (IDFT). Hence, the high-frequency information is suppressed when the value of frequency exceeds the cut-off frequency.

In order to reduce the calculation, we convert the Gaussian low-pass filter from the frequency domain to the spatial domain. The function of the digital spatial Gaussian low-pass filter $\bm G(\cdot)$ is defined as Eq (\ref{butt}), where $ -\lfloor m/2 \rfloor \leq x \leq \lfloor m/2 \rfloor$, $-\lfloor m/2 \rfloor \leq y \leq \lfloor m/2 \rfloor$. 
\begin{equation}
    G\left(x,y\right) = \frac{1}{2 \pi \lfloor m/2 \rfloor^2} e^{-(x^2+y^2)/(2\lfloor m/2 \rfloor^2) }  \label{butt}
\end{equation}

\subsubsection{The way to utilize the LFM}

\begin{figure}
    \small
    \centering
    %\vspace{-3mm}
    \includegraphics[width=1.0\linewidth]{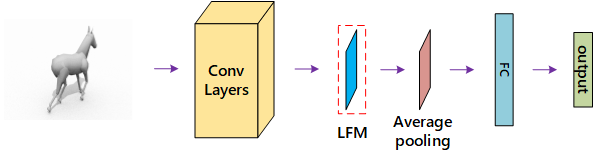}
    \vspace{-5mm}
    \caption{An overview of the IE domain adaptation model. IE: Insert the end of network. LFM: Low-frequency module.}
    \label{lfm}
    \vspace{-3mm}
\end{figure}

The LFM operates on the feature maps to obtain the low-frequency information of feature maps. There are two ways to utilize the LFM in the network:

\noindent \textbf{Insert the end of network (IE)}. We insert the LFM before the global average pooling layer to extract low-frequency information contained in feature maps as shown in Fig~\ref{lfm}. This design can ensure that the feature maps processed by the linear classification layer are the low-frequency information. The IE optimization method of the network is given with Eq (\ref{eq2},\ref{eq22}). 
\begin{equation}
    \hat{\theta_f}, \hat{\theta_g} = {\rm arg min}\ F(\theta_f, \theta_g ) 
\label{eq2}
\end{equation}
\begin{equation}
    F(\theta_f, \theta_g ) =  L\left(g\left( p \left(LFM \left(f\left(X_S, \theta_f\right), \theta_g\right) \right) \right),Y_S\right) 
\label{eq22}
\end{equation}

\begin{figure}
    
    \centering
    %\vspace{-2mm}
    \includegraphics[width=0.4\linewidth]{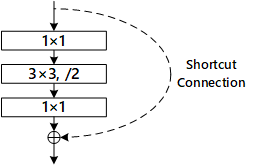}
    \includegraphics[width=0.4\linewidth]{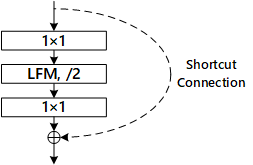}
    \vspace{-2mm}
    \caption{Visualization of the normal and RSL-equipped bottleneck. \textbf{Left} consists of $3\times3$ strided-convolutions and $1\times1$ convolutions. \textbf{Right} consists of the RSL and $1\times1$ convolution structure. RSL: Replace strided-convolution layers.}
    \label{rsl}
    \vspace{-5mm}
\end{figure}

\noindent \textbf{Replace strided-convolution layers (RSL)}. The down-sampling operation of CNNs can extract low-frequency information because the operation can result in reducing the size of feature map. Nevertheless, it is unstable and prone to lose crucial information since the operation does not obey the Nyquist Theorem according to \cite{zhang2019making}. Hence, we replace strided-convolution layers with the LFM in the encoder network as shown in Fig~\ref{rsl}. Different from strided-convolution, the LFM performs the low-pass filtering operation on each input feature map and its parameters are fixed.

\subsection{Combined with Other Methods}

Different  from  the  existing UDA methods,  our  method deals  with  the   domain adaptation  problem  from  the  perspective of frequency. Hence, our method is different from other methods in dealing with problems.  Our method is formulated as a plug-and-play unit that can be used to combine with existing UDA methods to achieve  better generalization performance. In the Experiments, we apply \cite{kang2019contrastive}, \cite{krumo_2019}, and \cite{xu2020exploring} to assist our method, and achieve state-of-the-art performance on multiple computer vision tasks.
\section{Experiments}

\subsection{Experimental Setups}

The $3\times 3$ convolution is the current popular structure. Hence, the $m$ of our LFM sets as 3 by default. To show the effectiveness of the proposed LFM, we first perform small image classification experiments for domain adaptation on the Office-31 \cite{saenko2010adapting} dataset to verify our method. On Office-31, similar to \cite{ganin2015unsupervised,long2015learning}, we validate the pairwise domain adaptation performance of our method on all six pairs of domains and take the average accuracy. Then we experiment with a challenging test-bed for UDA with the domain shift from synthetic data to real imagery on VisDA-2017 \cite{peng2017visda}. On VisDA-2017, we follow the full protocol \cite{kang2019contrastive} for the training setting but $D_0$ is set as 0.85, unlike the original 1.0. Because $D_0$ represents the cluster limit threshold, our method brings the same classes closer, and the threshold setting should be stricter. To explore LFM's generality further, we also conduct multi-label object detection experiments from Cityscapes \cite{cordts2016the} to FoggyCityscapes \cite{sakaridis2018semantic}, and we follow these two settings \cite{xu2020exploring} and \cite{krumo_2019} and fine-tune the network for adaptation experiments from Cityscapes to FoggyCityscapes. All models are trained from scratch on NVIDIA V100 GPUs with the default data augmentation and training strategy which are optimized for the vanilla model and no other tricks are used.

\setlength{\tabcolsep}{3pt}
\begin{table}
\caption{Results of the different strategies. The mean accuracy over six tasks on Office-31 is reported based on ResNet-50 \cite{he2016deep}. Our methods are trained with Gaussian high-pass pre-processing images, Gaussian low-pass pre-processing images, insert the end of network and replace strided-convolution layers, respectively.}
\vspace{-4mm}
\begin{center}
\scalebox{0.76}{
\begin{tabular}{ l  c  c cc c}
\toprule
Dataset                     & High-pass Pre-process &  Low-pass Pre-process           &  IE     & RSL   & Average\\
\midrule
\multirow{5}{*}{Office-31}  & &               &          &   & 76.1 \\
                            & $\checkmark$ &  &          &   & 73.2 \\    
                            & &  $\checkmark$ &              &   & 78.0 \\
                            & &               & $\checkmark$ &   & 81.4 \\
                            & &         &           &$\checkmark$& \textbf{81.6} \\
\bottomrule
\end{tabular}
}
\end{center}

\label{tab:low-frequency}
\vspace{-5mm}
\end{table}

\subsection{Ablation Studies}

\subsubsection{Effect of the different frequency components} 
\label{low-exp}
The \emph{Source-finetune} is the baseline method for cross-domain task. Hence, we first test the result of source-finetune on Office-31 dataset. As shown in the Table \ref{tab:low-frequency}, the first line shows the result of baseline method (76.1). 

To validate the effect of high-frequency information for cross-domain problem, we adopt Gaussian high-pass filtering to pre-process the Office-31 datasets. The result reveals that the high-frequency information of images limits the generalization ability of the model (from 76.1 to 73.2) in the second line. It is reasonable that the high-frequency information of image data contains domain-related information. Furthermore, we utilize Gaussian low-pass filtering to pre-process the Office-31 datasets to verify the effectiveness of low-frequency information. It can be observed that the result is better than the source-finetune (from 76.1 to 78.0), which means the low-frequency information is beneficial to alleviate domain adaptation task. From above results, it proves that our assumption that  the low-frequency  information  is more domain-invariant while the high-frequency information contains  domain-related  information. 

\begin{figure}
    \small
    \centering
    
    \includegraphics[width=0.4\linewidth]{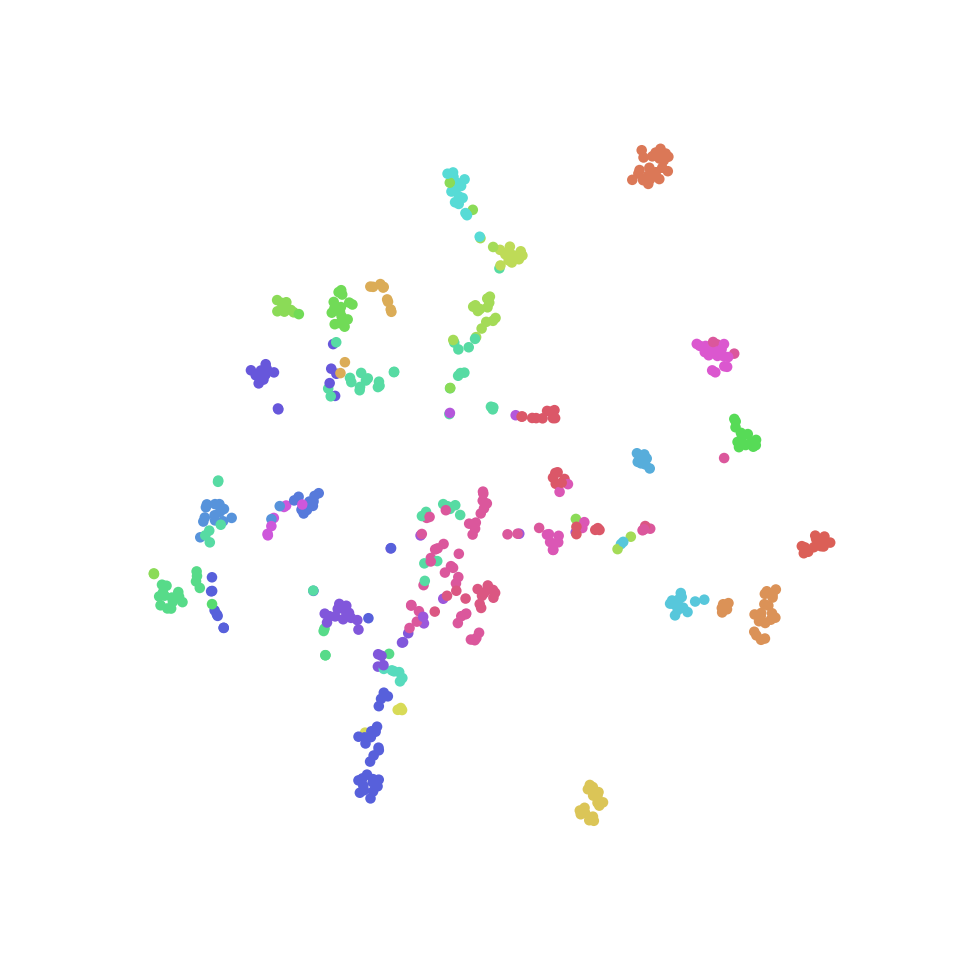}
    \includegraphics[width=0.4\linewidth]{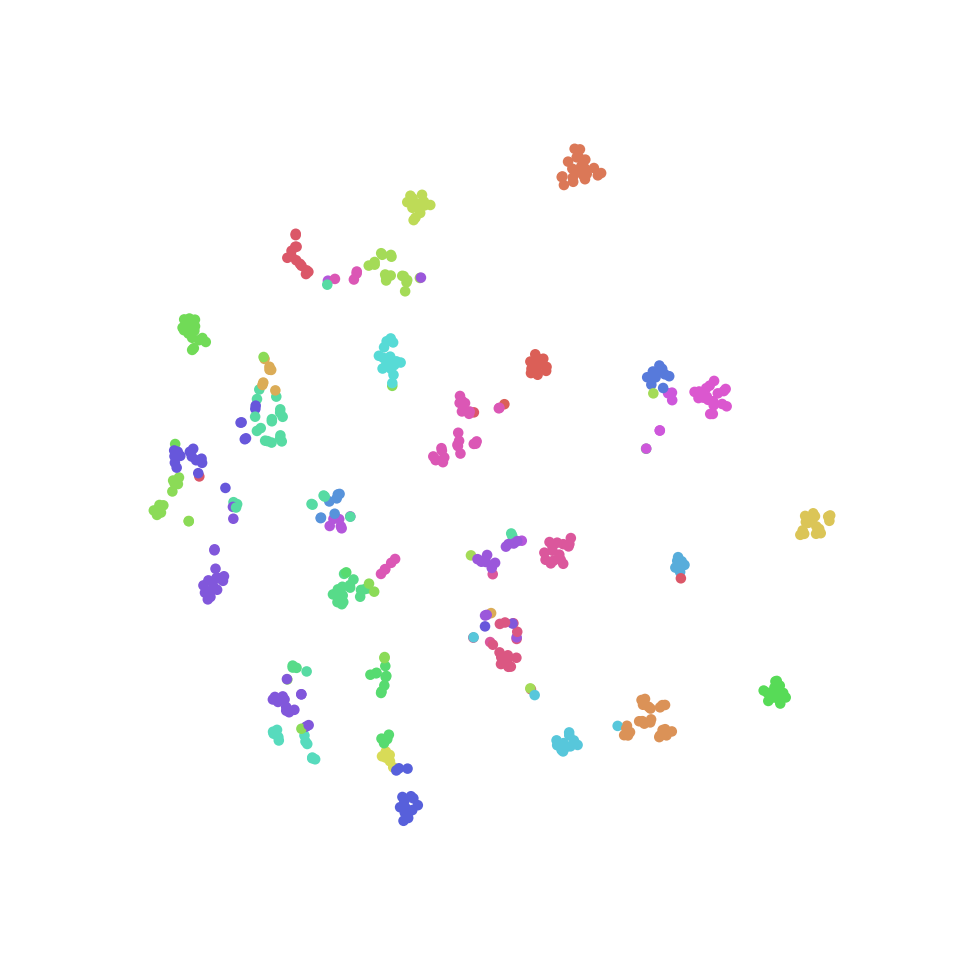}
    \vspace{-3mm}
    
    \caption{Visualization with t-SNE for different methods. \textbf{Left}: t-SNE of \emph{source-finetune}. \textbf{Right}: IE. The input activations of the last fully-connected layer are used for the computation of t-SNE. The results are Office-31 task \textbf{A $\rightarrow$ D}. The same color represents the same class while different color means different category.}
    \label{tsneda}
    \vspace{-3mm}
\end{figure}

\subsubsection{Effect of the LFM method}
It is noted that we adopt the IE strategy to train models to further utilize the low-frequency information of the dataset, as shown in Fig~\ref{tab:low-frequency}. It can be observed that introducing the IE further improves the adaptation performance and is better than the operation of Gaussian low-pass filtering (from 78.0 to 81.4). This phenomenon shows that it is better for the network to adaptively extract low-frequency features than the pre-processing dataset. Finally, we also adopt the RSL strategy to train models, and the result show impressive performance and are sightly better than IE (from 81.4 to 81.6). The two results demonstrate the effectiveness of our designs. Meanwhile, we visualize the distribution of learned features by t-SNE. As shown in Fig.~\ref{tsneda}, it illustrates a representative task \textbf{A $\rightarrow$ D}. Compared to source-finetune, the target feature representations learned by IE demonstrate higher intra-class compactness and a much larger inter-class margin. This suggests that utilizing low-frequency information can extract features that are invariant to different domains.

\subsubsection{Compared with other methods}
The LFM method is compared with two existing basic mainstream methods in the UDA field: RevGrad \cite{ganin2015unsupervised}, and DAN \cite{long2015learning}, to verify the merit of the proposed LFM. As Table \ref{table:cls-office-31} shows, both the IE and RSL methods are better than the DAN method, and the RevGrad method is slightly better than the IE and RSL methods. It should be emphasized that our method alleviates the cross-domain problem from the perspective of frequency, and it is different from the existing methods. Therefore, our method is orthogonal and complementary to the existing methods. For verifying the conjunction of our LFM method, we adopt the DAN and RevGrad to assist our method, respectively. 

Firstly, we utilize the DAN method to assist the IE and the final result is better than the DAN (from 80.4 to 82.3). Then, we also construct the RSL experiment and its result is also better than that of DAN (from 80.4 to 82.3). The performance of \emph{RSL+DAN} is equivalent to that of \emph{IE+DAN}. Although the performance of IE and RSL is better than that of DAN, the performance of the combination of the LFM method and DAN outperforms the single method.

For the RevGrad method, we apply it to assist the IE method and its performance is better than that of RevGrad (from 82.2 to 83.1). Similarly, the result of RSL is also better than the RevGrad (from 82.2 to 83.2). Meanwhile, the result of \emph{RSL+RevGrad} is slightly better than that of \emph{IE+RevGrad} (from 83.1 to 83.2). 

These results show the effect of the LFM method for alleviating the domain adaptation problem and prove our low-frequency assumption.

\subsection{Comparison with the State-of-the-art}
For fair comparison, we adopt the same backbone and re-implement them. Our re-implementations achieve comparable performance compared to original papers.
\subsubsection{Classification results}
VisDA is a challenging testbed for UDA with the domain shift from synthetic data to real imagery.  In total there are $\sim$280k images from 12 categories. The images are split into three sets, a training set with 152,397 synthetic images, a validation set with 55,388 real-world images, and a test set with 72,372 real-world images.
As shown in Table \ref{visda}, the \emph{Average} indicates the classification accuracy of 12 classes on VisDA-2017 with the validation set as the target domain by utilizing different UDA methods. Our method outperforms the popular UDA methods: RevGrad, DAN, self-ensembling (SE) (the first place in VisDA-2017 competition), and CAN. The mean accuracy of our RSL method outperforms that of the current state-of-the-art method CAN by 0.5 (from 86.8 to 87.3) on the VisDA-2017 validation dataset and the \emph{IE+CAN} method is slightly better than the \emph{RSL+CAN} method (from 87.3 to 87.4). On such a large dataset, the results reveal the potential and effectiveness of LFM.

\begin{table}[t]
%\begin{spacing}{1.1}
\caption{Classification accuracy (\%) for all the six tasks of Office-31 dataset based on ResNet-50 \cite{he2016deep}.}
\vspace{-2mm}
\centering
\small
\setlength{\tabcolsep}{1.0mm}
\scalebox{0.58}{
\begin{tabular}{ l  c  c   c   c   c   c  c}
\toprule
Method &  A $\rightarrow$ W & D $\rightarrow$ W & W $\rightarrow$ D & A $\rightarrow$ D & D $\rightarrow$ A & W $\rightarrow$ A & Average \\
\midrule
Source-finetune & 68.4 $\pm$ 0.2 & 96.7 $\pm$ 0.1 & 99.3 $\pm$ 0.1 & 68.9 $\pm$ 0.2 & 62.5 $\pm$ 0.3 & 60.7 $\pm$ 0.3 & 76.1 \\
%RTN \cite{long2016unsupervised}   & 84.5 $\pm$ 0.2 & 96.8 $\pm$ 0.1 & 99.4 $\pm$ 0.1 & 77.5 $\pm$ 0.3 & 66.2 $\pm$ 0.2 & 64.8 $\pm$ 0.3 & 81.6 \\ 
DAN \cite{long2015learning}   & 80.5 $\pm$ 0.4 & 97.1 $\pm$ 0.2 & 99.6 $\pm$ 0.1 & 78.6 $\pm$ 0.2 & 63.6 $\pm$ 0.3 & 62.8 $\pm$ 0.2 & 80.4 \\
RevGrad \cite{ganin2015unsupervised} & 82.0 $\pm$ 0.4 & 96.9 $\pm$ 0.2 & 99.1 $\pm$ 0.1 & 79.7 $\pm$ 0.4 & 68.2 $\pm$ 0.4& 67.4 $\pm$ 0.5 & 82.2 \\ 
\midrule
Ours (IE+Source-finetune) & 77.3 $\pm$ 0.2  &  96.7 $\pm$ 0.2   &  \textbf{99.8 $\pm$ 0.2} &  83.0 $\pm$ 0.2 & 65.8 $\pm$ 0.2 & 65.6 $\pm$ 0.2 & 81.4      \\
Ours (RSL+Source-finetune) & 77.5 $\pm$ 0.2  &  97.0 $\pm$ 0.2   &  \textbf{99.8 $\pm$ 0.2} &  83.2 $\pm$ 0.2 & 66.2 $\pm$ 0.2 & 66.0 $\pm$ 0.2 & 81.6      \\
Ours (IE+DAN) & 80.3 $\pm$ 0.2  &  97.0 $\pm$ 0.2   &  \textbf{99.8 $\pm$ 0.2} &  \textbf{83.4 $\pm$ 0.2} & 66.8 $\pm$ 0.2 & 66.3 $\pm$ 0.2 & 82.3      \\
Ours (RSL+DAN) & 80.4 $\pm$ 0.2  &  97.1 $\pm$ 0.2   &  \textbf{99.8 $\pm$ 0.2} &  83.2 $\pm$ 0.2 & 67.0 $\pm$ 0.2 & 66.0 $\pm$ 0.2 & 82.3      \\
Ours (IE+RevGrad) & \textbf{82.6 $\pm$ 0.3}  &  96.9 $\pm$ 0.2   &  \textbf{99.8 $\pm$ 0.2} &  82.8 $\pm$ 0.4 & 68.8 $\pm$ 0.3 & \textbf{68.0 $\pm$ 0.4} & 83.1      \\
Ours (RSL+RevGrad) & 82.5 $\pm$ 0.2  & \textbf{ 97.3 $\pm$ 0.2 }  &  \textbf{99.8 $\pm$ 0.2} &  83.1 $\pm$ 0.4 & \textbf{69.1 $\pm$ 0.3} & 67.5 $\pm$ 0.4 & \textbf{83.2}      \\
%Ours (JCAN)     & 91.6 $\pm$ 0.3  & \textbf{98.3} $\pm$ 0.2 & \textbf{99.9} $\pm$ 0.1 & \textbf{89.2} $\pm$ 0.3  &  \textbf{74.3} $\pm$ 0.2 &  \textbf{74.8} $\pm$  0.1 & \textbf{88.0}   \\
\bottomrule
\end{tabular}
%\end{spacing}
%\vspace{-1mm}
}
\label{table:cls-office-31}
\vspace{-3mm}
\end{table}

\begin{table}[t]
\caption{Classification accuracy (\%) on the VisDA-2017 validation set based on ResNet-101 \cite{he2016deep}.}
\vspace{-2mm}
%\begin{spacing}{1.1}
\centering
\small
\setlength{\tabcolsep}{1.0mm}
\scalebox{0.65}{
\begin{tabular}{ l  c  c   c   c   c   c  c c c c c c c}
\toprule
Method & \rotatebox{90}{airplane} & \rotatebox{90}{bicycle} & \rotatebox{90}{bus} & \rotatebox{90}{car} & \rotatebox{90}{horse} & \rotatebox{90}{knife} & \rotatebox{90}{motorcycle} & \rotatebox{90}{person} & \rotatebox{90}{plant} & \rotatebox{90}{skateboard} & \rotatebox{90}{train} & \rotatebox{90}{truck} & Average \\
\midrule
Source-finetune   &  72.3 & 6.1 & 63.4 & \textbf{91.7} & 52.7 & 7.9 & 80.1 & 5.6  & 90.1 & 18.5 & 78.1 & 25.9 & 49.4 \\
RevGrad \cite{ganin2015unsupervised} & 81.9 & 77.7 & 82.8 & 44.3 & 81.2 & 29.5 & 65.1 & 28.6 & 51.9 & 54.6 & 82.8 & 7.8 & 57.4 \\
DAN \cite{long2015learning}   & 68.1 & 15.4 & 76.5 & 87.0 & 71.1 & 48.9 & 82.3 & 51.5 & 88.7 & 33.2 & \textbf{88.9} & 42.2 & 62.8 \\
JAN \cite{long2017deep}       & 75.7 & 18.7 & 82.3 & 86.3 & 70.2 & 56.9 & 80.5 & 53.8 & 92.5 & 32.2 & 84.5 & 54.5 & 65.7 \\
%AD^{3} \cite{yang2021robust} & - & - & - & - & - & - & - & - & - & - & - & - & 68.5 \\
%MCD \cite{saito2017maximum}   & 87.0 & 60.9 & 83.7 & 64.0 & 88.9 & 79.6 & 84.7 & 76.9 & 88.6 & 40.3 & 83.0 & 25.8 & 71.9 \\
%ADR \cite{saito2017adversarial}   & 87.8 & 79.5 & 83.7 & 65.3 & 92.3 & 61.8 & 88.9 & 73.2 & 87.8 & 60.0 & 85.5 & 32.3 & 74.8 \\
%DTA \cite{lee2019drop} &  93.7 & 82.2   & 85.6 &  83.8  & 93.0  &  81.0  & 90.7 & 82.1 & 95.1 & 78.1 & 86.4 & 32.1 & 81.5 \\
GSDA \cite{hu2020unsupervised} &  93.1 & 67.8   & 83.1 &  83.4  & 94.7  &  93.4  & 93.4 & 79.5 & 93.0 & 88.8 & 83.4 & 36.7 & 81.5 \\

SE \cite{french2018self}  &  95.9 & 87.4   & 85.2 &  58.6  & 96.2  &  95.7  & 90.6 & 80.0 & 94.8 & 90.8 & 88.4 & 47.9 & 84.3 \\

CAN  \cite{kang2019contrastive}    & 96.7 & \textbf{90.3} & 84.2 & 66.4 & 96.5 & 97.1 & 88.0 & 83.0 & 96.1 & 95.0 & 87.0 & \textbf{61.3} & 86.8 \\
\midrule
%Ours (inital clustering) & \textbf{97.9} & 87.7 & 80.8 & 72.2 & 93.2 & 75.8 & 77.4 & 57.9 & 92.1 & 43.2 & 93.9 & 68.7 & 78.4\\
Ours (RSL+CAN)     & \textbf{97.5} & 86.1 & 84.7 & 71.7 & \textbf{96.2} & \textbf{98.2} & 90.6 & 82.7 & 96.8 & 94.8 & \textbf{88.9}& 59.5 & 87.3 \\
Ours (IE+CAN)     & 96.8 & 85.8 & \textbf{85.3} & 72.8 & 95.8 & 97.3 & \textbf{91.7} & \textbf{84.0} & \textbf{97.3} & \textbf{95.1} & 87.1 & 59.8 & \textbf{87.4} \\
\bottomrule
\end{tabular}
%\end{spacing}
%\vspace{-1mm}
}
\label{visda}
\vspace{-5mm}
\end{table}

\begin{table}[t]
\caption{Results (\%) on adaptation from Cityscapes to Foggy-Cityscapes (normal $\to$ foggy). The backbone network is ResNet-50.} \label{cityres}
\vspace{-2mm}
%\begin{spacing}{1.1}
\centering
\small
\setlength{\tabcolsep}{1.0mm}
\scalebox{0.75}{
\begin{tabular}{lccccccccc}
\toprule[1.0pt]
Methods & person & rider & car & truck & bus & train & motorcycle & bicycle & mAP \\
\hline

Source-only & 26.9 & 38.2 & 35.6 & 18.3 & 32.4 & 9.6 & 25.8 & 28.6 & 26.9 \\

%DivMatch \cite{diversify_and_match} & 31.8 & 40.5 & 51.0 & 20.9 & 41.8 & 34.3 & 26.6 & 32.4 & 34.9 \\
%MTOR \cite{zhu2019adapting} & 30.6 & 41.4 & 44.0 & 21.9 & 38.6 & 40.6 & 28.3 & 35.6 & 35.1 \\
%SW-DA-Faster \cite{saito2019strong} & 31.8 & 44.3 & 48.9 & 21.0 & 43.8 & 28.0 & 28.9 & 35.8 & 35.3 \\
SC-DA-Faster \cite{zhu2019adapting} & 33.8 & 42.1 & 52.1 & 26.8 & 42.5 & 26.5 & 29.2 & 34.5 & 35.9 \\
GPA \cite{xu2020cross} & 32.9 & 46.7 & 54.1 & 24.7 & 45.7 & 41.1 & 32.4 & 38.7 & 39.5 \\
KR-DA-Faster \cite{krumo_2019} & 36.8 & 46.4 & 54.5 & 27.7 & 47.3 & 42.7 & 32.7 & 38.6 & 40.8 \\
\hline
Our (IE+KR)  & 36.8 & 46.9 & 52.9 & \textbf{28.9} & 48.2 & 47.1 & 31.7 & 38.9 & 41.4  \\
Our (RSL+KR) & \textbf{37.1} & \textbf{47.6} & \textbf{55.0} & 28.3 & \textbf{48.5} & \textbf{47.8} & \textbf{32.8} & \textbf{39.8} & \textbf{42.1} \\
% GPA (Two-stage Align) & 32.4 & \textbf{44.8} & \textbf{54.6} & 25.1 & \textbf{47.1} & 34.0 & \textbf{33.5} & 34.4 & \textbf{38.2} \\

\bottomrule[1.0pt]
\end{tabular}
%\end{spacing}
%\vspace{-1mm}
}
\vspace{-4mm}
\end{table}

\subsubsection{Object detection results}
To verify the generality of our method, we construct object detection experiments, and adopt current state-of-the-art methods: DA-Faster-ICR-CCR \cite{xu2020exploring} and KR-DA-Faster \cite{krumo_2019} as our baseline methods.

First, we train our method with the state-of-the-art method KR-DA-Faster.
Its backbone is widely popular ResNet-50 and the network initializes with Caffe pre-trained weights. Ultimately, the model achieves the best performance thus far from Cityscapes to Foggy-Cityscapes by adopting our method. Table \ref{cityres} shows the comparison results. Our \emph{IE+KR} can boost the performance of KR-DA-Faster by 0.6 mAP (from 40.8 to 41.4). The \emph{RSL+KR} method adopts Pytorch pre-trained weights and still outperforms the KR-DA-Faster by 1.3 mAP (from 40.8 to 42.1) although Caffe pre-trained models have better performance than Pytorch pre-trained. In particular, our RSL can greatly improve the detection results in the target domain. The results reveal that the RSL is better than IE in the object detection task. In particular, our method can greatly improve the detection results for some difficult categories such as “train”. The RSL method outperforms the state-of-the-art by 5.1 mAP for the training class. This clearly verifies the importance of low-frequency information for cross-domain object detection. 

\begin{table}[t]
\caption{Results (\%) on adaptation from Cityscapes to Foggy-Cityscapes (normal $\to$ foggy). The backbone network is VGG-16.} \label{cityvgg}
\vspace{-2mm}
	%\begin{spacing}{1.1}
		\centering
		\small
		\scalebox{0.65}{
		\setlength{\tabcolsep}{1.0mm}
		\begin{tabular}{lccccccccc}
			\toprule[1.0pt]
			Method      & persn & rider & car & truck & bus & train & mbike & bcycle & mAP \\
			\midrule
			Source Only  & 24.1             & 33.1          & 34.3          & 4.1           & 22.3  & 3.0   & 15.3  & 26.5  & 20.3  \\
			DA-Faster \cite{chen2018domain}       & 25.0          & 31.0          & 40.5          & 22.1  & 35.3  & 20.2  & 20.0  & 27.1  & 27.6  \\
			
			DA-Faster-ICR-CCR \cite{xu2020exploring}   & 29.7 & 37.3          & \textbf{43.6}          & 20.8   & \textbf{37.3}  & 12.8  & \textbf{25.7}  & 31.7  & 29.9  \\
			\midrule
			Our method (RSL+DA-Faster-ICR-CCR)      & \textbf{30.1}           & \textbf{42.9}         & 43.3 & \textbf{24.1} & 35.2 & \textbf{20.5} & 25.4  & \textbf{33.6}  & \textbf{31.9} \\
			\bottomrule[1.0pt]
		\end{tabular}
		}
	%\end{spacing}
%\vspace{-1mm}

\vspace{-3mm}
\end{table}

Moreover, DA-Faster-ICR-CCR is an extension method based on DA-Faster. Its backbone is VGG-16. We combine our RSL idea with VGG-16 and apply it to the DA-Faster-ICR-CCR method. As shown in Table \ref{cityvgg}, we observe that our method outperforms DA-Faster-ICR-CCR by 2.0 mAP (from 29.9 to 31.9). The result demonstrates that our method is compatible with other backbones.

\section{Conclusion}
\label{conclusions}
In this paper, we propose an assumption that low-frequency information is more domain-invariant and more suitable for domain adaptation tasks while the high-frequency information contains domain-related information in different domains. Meanwhile, we construct massive experiments and visualization analysis to demonstrate the assumption. Finally, we introduce a method, named LFM, to combine with existing UDA methods easily and achieve better performance. Our method outperforms state-of-the-art methods on VisDA-2017 and Cityscapes to FoggyCityscapes. In future, we will introduce RSL to the current self-supervised methods \cite{li2021mst,li2022univip} we have already explored.

\noindent \textbf{Acknowledgement.} This work was supported by Key-Area Research and Development Program of Guangdong Province (No.2021B0101410003), National Natural Science Foundation of China under Grants No.62002357, No.62176254, No.61976210, No.61876086,  No.62076235 and No.62006230.

\small
\bibliographystyle{IEEEbib}
\bibliography{icme2022template}

\end{document}